\documentclass[sigconf]{acmart}

\AtBeginDocument{%
  }

\usepackage[utf8]{inputenc} % allow utf-8 input
\usepackage[T1]{fontenc}    % use 8-bit T1 fonts
\usepackage{hyperref}       % hyperlinks
\usepackage{url}            % simple URL typesetting
\usepackage{booktabs}       % professional-quality tables
\usepackage{amsfonts}       % blackboard math symbols
\usepackage{nicefrac}       % compact symbols for 1/2, etc.
\usepackage{microtype}      % microtypography
\usepackage{xcolor}         % colors

\usepackage{multirow}
\usepackage{subfigure}

\usepackage[english]{babel}
\usepackage{moresize}
\usepackage{amsmath}
\usepackage{algorithmic}
\usepackage{balance}
\usepackage{comment}
\usepackage{paralist}
\usepackage{bm}
\usepackage{pgfplots}
\usetikzlibrary{pgfplots.dateplot}

\usepackage{flushend}
\usepackage[english]{babel}
\usepackage{graphicx}

\usepackage{amssymb}
\usepackage{amsfonts}
\usepackage{url}
\usepackage{bbm}
\usepackage{longtable}
\usepackage{rotating}
\usepackage{multirow}
\usepackage{mathrsfs}
\usepackage{enumitem}
\usepackage[linesnumbered,algoruled,boxed,lined]{algorithm2e}
\usepackage{adjustbox}
\usepackage{hyperref}
\usepackage{pgfplots}
\usetikzlibrary{pgfplots.dateplot}
\usepackage{filecontents}

\newcommand{\eg}{\textit{e}.\textit{g}.}

\def\model{AnyGraph}

\newcommand{\graph}{\mathcal{G}}
\newcommand{\setv}{\mathcal{V}}
\newcommand{\sete}{\mathcal{E}}
\newcommand{\setm}{\mathcal{M}}
\newcommand{\sety}{\mathcal{Y}}

\newcommand{\sets}{\mathcal{S}}

\newcommand{\matf}{\textbf{F}}
\newcommand{\dmnr}{\mathbb{R}}

\newcommand{\vecb}{\textbf{b}}
\newcommand{\vecf}{\textbf{f}}
\newcommand{\vece}{\textbf{e}}

\newcommand{\mata}{\textbf{A}}

\newcommand{\mate}{\textbf{E}}
\newcommand{\matu}{\textbf{U}}
\newcommand{\matv}{\textbf{V}}
\newcommand{\matw}{\textbf{W}}
\newcommand{\loss}{\mathcal{L}}
\newcommand{\param}{\mathbf{\Theta}}

\setcopyright{none}

\acmConference[ACM KDD]{}{Aug 3–Aug 7, 2025}{Woodstock, NY}

\begin{document}

%%
%% The "title" command has an optional parameter,
%% allowing the author to define a "short title" to be used in page headers.
% \title{Automated Mixture-of-Experts for Graph Foundation Models}
\title{AnyGraph: Graph Foundation Model in the Wild}

\author{Lianghao Xia and Chao Huang$^*$}
\thanks{$*$ Chao Huang is the Corresponding Author.}
\affiliation{
    \institution{The University of Hong Kong}
    \country{\{lhaoxia, chuang7\}@hku.hk}\\
    \country{\textbf{Github}: \href{https://github.com/HKUDS/AnyGraph}{https://github.com/HKUDS/AnyGraph}}
}
% \affiliation{
%     \institution{The University of Hong Kong}
%     % \country{Hong Kong SAR}
% }
% \affiliation{
%     \institution{\{lhaoxia, chuang7\}@hku.hk}
% }
% \affiliation{
%     \institution{\textbf{Github}: \href{https://github.com/HKUDS/AnyGraph}{https://github.com/HKUDS/AnyGraph}}
%     \country{}
% }

\begin{abstract}
The growing ubiquity of relational data structured as graphs has underscored the need for graph learning models with exceptional generalization capabilities. However, current approaches often struggle to effectively extract generalizable insights, frequently requiring extensive fine-tuning and limiting their versatility. Graph foundation models offer a transformative solution, with the potential to learn robust, generalizable representations from graph data. This enables more effective and adaptable applications across a wide spectrum of tasks and domains. In this work, we investigate a unified graph model, \model, designed to handle key challenges: i) \textbf{Structure Heterogenity}. Addressing distribution shift in graph structural information; ii) \textbf{Feature Heterogenity}. Handling diverse feature representation spaces across graph datasets; iii) \textbf{Fast Adaptation}. Efficiently adapting the model to new graph domains; iv) \textbf{Scaling Law Emergence}. Enabling the model to exhibit scaling law behavior, where its performance scales favorably with the amount of data and parameter sizes. To tackle these critical challenges, we build the \model\ upon a Graph Mixture-of-Experts (MoE) architecture. This approach empowers the model to effectively manage both the in-domain and cross-domain distribution shift concerning structure-level and feature-level heterogeneity. Furthermore, a lightweight graph expert routing mechanism is proposed to facilitate \model's fast adaptability to new data and domains. Our extensive experiments on diverse 38 graph datasets have demonstrated the strong zero-shot learning performance of \model\ across diverse graph domains with significant distribution shift. Furthermore, we have validated the model's fast adaptation ability and scaling law emergence, showcasing its versatility. 
% We have anonymously released our open-sourced \model\ implementation at the following link: {\color{blue}\url{https://anonymous.4open.science/r/\model-FECD}}.
\end{abstract}

% \keywords{Do, Not, Us, This, Code, Put, the, Correct, Terms, for, Your, Paper}

\maketitle

\section{Introduction}
\label{sec:intro}

% \begin{figure}
%     \centering
%     \subfigure[Generalizability of \model.]{
%         \includegraphics[width=0.43\columnwidth]{figs/intro1.pdf}
%     }
%     \subfigure[Scaling law of \model]{
%         \includegraphics[width=0.47\columnwidth]{figs/param_zeroshot_multidomain.pdf}
%     }
%     % \subfigure[]{
%     %     \includegraphics[width=0.31\columnwidth]{figs/data_zeroshot_multidomain.pdf}
%     % }
%     \vspace{-0.15in}
%     \caption{The generalizability and scaling law of \model.}
%     \label{fig:enter-label}
% \end{figure}

The growing ubiquity of relational data in the form of graphs has underscored the pressing need for advanced graph learning models that excel at generalization~\cite{feyposition,jin2020graph}. As real-world applications of graph-structured data continue to proliferate across diverse domains, including social networks, academic networks, transportation systems, and biological networks, the ability of graph learning models to effectively handle distribution shifts and adapt to new graph domains has become increasingly crucial~\cite{zhang2023cross,zhao2024all,xia2023automated,mao2024demystifying}. Developing models with robust zero-shot learning performance and fast adaptation capabilities can unlock transformative opportunities for leveraging the rich insights encoded within graph data.

The field of graph learning has seen significant advancements in recent years, largely driven by the power of Graph Neural Networks (GNNs)~\cite{liu2022graph,xiao2021learning,li2021training}. However, the current state-of-the-art models often fall short when it comes to truly generalizable performance. Existing approaches tend to be heavily reliant on arduous fine-tuning processes, making them ill-equipped to handle the diverse array of graph structures and distributions encountered in real-world applications. This inability to adapt swiftly and seamlessly to novel graph domains poses a critical barrier to the widespread adoption of graph learning technologies. Therefore, addressing this challenge is of paramount importance if we are to fully harness the transformative potential of graph-based insights.

Inspired by the principles that have driven the development of successful foundation models in understanding vision and language data~\cite{wang2022omnivl,wang2023internimage}, the concept of a versatile graph foundation model holds immense potential to unlock new frontiers in graph learning. By learning rich, transferable representations from diverse graph-structured data, such a model can be efficiently adapted to a wide array of graph domains and tasks. However, building an effective and adaptive graph foundation model is not a trivial endeavor. Several key challenges must be overcome, including: \\\vspace{-0.12in}

\noindent (i) \textbf{Structure Heterogeneity}. The development of versatile graph models faces the challenge of accommodating diverse structural properties and data distributions in various graph datasets. For instance, graphs can exhibit substantial heterogeneity in node degree distributions, ranging from homogeneous to highly skewed patterns. Similarly, graph structures can vary greatly in complexity, from simple topologies to intricate, hierarchical arrangements. These structural variations can significantly impact the performance and generalization of graph learning algorithms. Effectively addressing this diversity is critical for developing unified models that can thrive across a wide range of graph-structured data.\\\vspace{-0.12in}

\noindent (ii) \textbf{Feature Heterogeneity}. Graphs exhibit substantial heterogeneity in their node and edge features, which can span categorical attributes, continuous numerical data, and multi-modal content. Furthermore, the dimensionality and semantics of these features often vary dramatically across different graph domains. For instance, a social interaction graph may include textual content and demographic information associated with its nodes, while a molecular graph may feature atomic compositions and bond types. Effectively handling this feature heterogeneity is crucial for building a versatile graph model capable of generalizing across diverse graph domains. \\\vspace{-0.12in}

\noindent (iii) \textbf{Fast Adaptation for Broad Applicability}. A key capability for effective graph foundation models is the ability to efficiently adapt to new graph dataset and domains. Rather than requiring extensive retraining or fine-tuning, the ideal model should be able to quickly adjust its parameters and learning strategies to handle the structural and distributional characteristics of previously unseen graph datasets. By seamlessly generalizing and performing well across a diverse range of real-world scenarios – from user behavior graphs to transportation networks and biological systems – these adaptable models can unlock transformative insights across an ever-expanding universe of graph-structured data. \\\vspace{-0.12in}

\noindent (iv) \textbf{Scaling Laws for Transformative Graph Capabilities}. A key characteristic of successful foundation models in domains like CV~\cite{cherti2023reproducible} and NLP~\cite{muennighoff2024scaling} is their ability to exhibit scaling laws - where performance systematically improves as the model size or training dataset increases. By harnessing this emergent scaling phenomenon, graph foundation models can unlock unprecedented levels of capability and generalization, far surpassing the limitations of fixed-capacity architectures. As the size of graph datasets and model complexity grow, these scaling-aware designs can continue delivering transformative performance gains. \\\vspace{-0.12in}

\begin{figure}[t]
    \centering
    \subfigure[Generalizability of \model.]{
        \includegraphics[width=0.42\columnwidth]{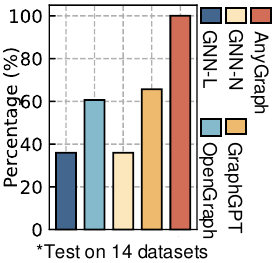}
    }
    \subfigure[Scaling law of \model]{
        \includegraphics[width=0.46\columnwidth]{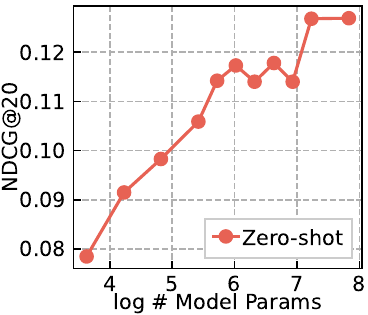}
    }
    % \subfigure[]{
    %     \includegraphics[width=0.31\columnwidth]{figs/data_zeroshot_multidomain.pdf}
    % }
    \vspace{-0.1in}
    \caption{\model's generalizability and scaling law reveals its exceptional capabilities. Compared to baseline methods, the superior performance of \model\ can be observed in its exceptional cross-domain generalization ability.}
    \label{fig:enter-label}
    \vspace{-0.1in}
\end{figure}

\noindent \textbf{The Presented Work}. To tackle the above challenges, our \model\ model is built upon a Mixture-of-Experts (MoE) architecture, which allows for effective handling of both the in-domain and cross-domain distribution shift in structure-level and feature-level heterogeneity. The proposed graph MoE paradigm empowers \model\ to learn a diverse ensemble of graph experts, each tailored to specific structural characteristics. This enables the model to effectively manage the distribution shift in graph topologies, ranging from homogeneous to highly skewed degree distributions, as well as handle graphs with varying levels of complexity. Furthermore, the MoE architecture of \model\ facilitates fast adaptation of the graph model. Rather than relying on a single, fixed-capacity model, the Graph MoE learns an ensemble of specialized expert networks, each tailored to capture distinct structural and feature-level characteristics of graph data. The lightweight graph expert routing mechanism allows \model\ to quickly identify and activate the most relevant experts for a given input graph, without requiring extensive retraining or fine-tuning across the entire model. The key findings of this work can be summarized as: \\\vspace{-0.12in}

\begin{itemize}[leftmargin=*]

\item \textbf{Methodology Design Motivations of \model}. Current large graph models~\cite{chen2024llaga,liuone,li2024zerog} often struggle when faced with the substantial heterogeneity found in real-world graph data. This is especially challenging when it comes to feature-level heterogeneity. These fixed-capacity models may encounter interference between different types of graph datasets, and can sometimes overfit to new data, leading to catastrophic forgetting. To address these challenges, the proposed Mixture-of-Experts (MoE) architecture for graph models was designed with a focus on adaptability. This new paradigm empowers the model to flexibly adjust to the nuances of diverse graph datasets, dynamically selecting the most appropriate experts to learn distinct patterns. \\\vspace{-0.12in}

\item \textbf{Stronger Gernealiation Capacities of \model}. Through our extensive experiments, the proposed \model\ model with the graph MoE framework has demonstrated strong generalization capacities across a wide range of graph tasks and domains. The experimental results showcase the \model's ability to outperform existing graph models in terms of both predictive performance and robustness to distribution shift. \\\vspace{-0.12in}

\item \textbf{Fast Adapability of \model}. Our innovative dynamic expert selection mechanism enhances \model's ability to swiftly adapt to new graph domains. By dynamically routing inputs through relevant experts, \model\ can quickly activate the specialized networks best suited for the task. This strong adaptation sets \model\ apart from baselines. Evaluation shows its superiority through rapid convergence and exceptional performance, further justifying its cross-domain versatility. \\\vspace{-0.12in}

\item \textbf{The Scaling Law of \model}. Our experiments reveal that \model's performance follows the scaling law, where the model continues to improve as model size and training data increase. Additionally, \model\ exhibits emergent abilities, where its generalization capabilities see sudden significant improvements with further scaling. This critical scaling law property has been largely overlooked in prior investigations, but it underscores the immense value that \model\ derives from its scaling-driven enhancements to generalization performance.

\end{itemize}

% \clearpage
\section{Preliminaries}
\label{sec:model}

\noindent\textbf{Graph-Structured Data}. A graph $\graph$ consists of a set of nodes $\setv = \{v_i\}$ and a set of edges $\sete = \{(v_i, v_j)\}$. In many cases, each node $v_i$ is associated with a feature vector $\vecf_i \in \dmnr^{d_0}$. To efficiently utilize such graph-structured data, the link information is typically recorded using an adjacency matrix $\mata \in \dmnr^{|\setv| \times |\setv|}$. Each element $a_{i,j}$ of $\mata$ is either 1 or 0, inddicating whether there is an edge from node $v_i$ to $v_j$. Additionally, the feature vectors of the nodes are usually represented by a feature matrix $\matf \in \dmnr^{|\setv| \times d_0}$, where each row corresponds to a node's feature vector. The primary goal of learning from such graph-structured data is to generate embeddings for the graph elements, typically nodes, that effectively capture both the structural and feature-based information of the graph.
\\\vspace{-0.08in}

\noindent\textbf{Graph Foundation Models (GFMs)}. 
% graph foundation models should possess the capabilities of cross-dataset prediction
The essence of GFMs lies in their strong generalization capabilities. Specifically, a graph foundation model should be able to handle unseen graph data that exhibits significant discrepancies from its training graph datasets. These discrepancies may include differences in feature spaces, as well as variations in node and edge semantics across datasets. Formally, let's denote the training graphs as $\mathbb{S} = {\graph_s}$, where each graph $\graph_s$ is associated with a label set $\sety_s$. Similarly, the set of test graphs is denoted as $\mathbb{T} = {\graph_t}$, with labels $\sety_t$. With a differentiable training objective $\loss$ and an evaluation criterion $\mathcal{C}$ to measure the prediction accuracy of downstream tasks, building a graph foundation model $f_\param$ with trainable parameters $\param$ can be formalized as follows:
\begin{align}
    \mathop{\arg\max}_{f, \mathcal{L}} \sum_{\mathcal{G}_t} \mathcal{C}\left(f_\param(\mathcal{G}_t), \mathcal{Y}_t\right),~\mathbf{\Theta} = \mathop{\arg\min}_{\mathbf{\Theta}} \sum_{\mathcal{G}_s}\mathcal{L}\left(f_\param(\mathcal{G}_s), \mathcal{Y}_s\right)
\end{align}
The above formulation reveals that the key to building graph foundation models are: \textbf{i)} the model architecture design ($f$), which must have the capacity to encode diverse feature spaces and structural patterns, and \textbf{ii)} the model training process ($\loss$), which must effectively traverse such diverse data to find an optimal solution $\param$ for the model $f$.
In light of this, our \model\ employs a mixture-of-experts architecture with an automated expert routing method, to seamlessly integrate powerful prediction models for highly diverse graph data. \model\ is extensively trained on graphs from various applications using multiple featuring methods, with a graph augmentation technique to further enhance data diversity.
\section{Methodology}
\label{sec:solution} 
\begin{figure*}
    \centering
    \includegraphics[width=0.95\textwidth]{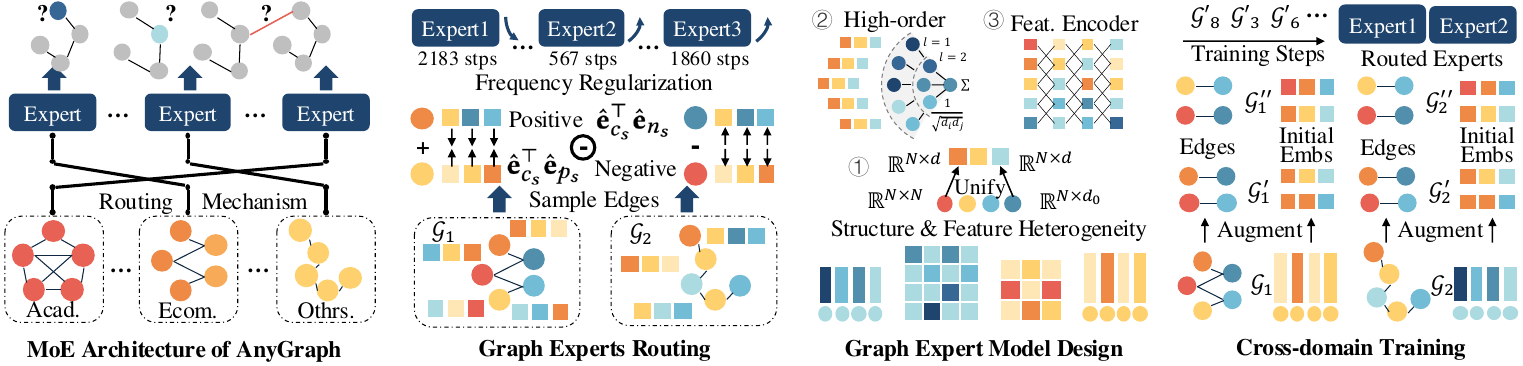}
    \vspace{-0.05in}
    \caption{The proposed graph Mixture-of-Experts (MoE) paradigm enables \model\ to learn a diverse ensemble of graph experts, each tailored to specific structural characteristics. The lightweight expert routing mechanism allows \model\ to quickly identify and activate the most relevant experts for a given input graph, without extensive retraining or fine-tuning.}
    \label{fig:framework}
    % \vspace{-0.05in}
\end{figure*}

The proposed \model\ framework aims to address both cross-domain and in-domain heterogeneity in graph structures and node features, while enabling fast adaptation to new data. The overall framework of \model\ is depicted in Fig.~\ref{fig:framework}.

\subsection{MoE Architecture of \model}
% Overall MoE Architecture
\subsubsection{\bf Addressing Cross-domain Graph Heterogeneity}
To model heterogeneous graph patterns across different domains, \model\ employs a MoE architecture that consists of multiple graph expert models, each responsible for handling graphs with specific characteristics. An automated routing algorithm is designed to assign input graph data to the most competent expert model for training and prediction.
Specifically, the \model\ framework can be denoted as $\setm = (f_{\param_1}, f_{\param_2}, \cdots, f_{\param_K}, \psi)$, where $K$ denotes the number of experts. For an input graph $\graph$, the routing algorithm $\psi$ firstly identifies the most competent expert model, and the corresponding model is then used for predicting the graph data, as:
\begin{align}
    \hat{y}_{i,j}=\hat{\vece}^{\top}_i \hat{\vece}_j, ~~~\hat{\mate} = f_{\param_k}(\graph), ~~~k=\psi(\graph)
\end{align}
where each expert model $f_{\param_k}$ can be viewed as a projection from the graph space to a node embedding space with uniquely trained parameters $\param_k$. And $\hat{y}_{i,j}$ represents the dot-product-based prediction of whether the entity $v_i$ should be related to the entity $v_j$. Here, $v_i$ and $v_j$ could be vanilla graph nodes, class labels, or graph labels.

\subsubsection{\bf Graph Expert Routing Mechanism}
% self-supervised loss as indicator
% training frequency regularizer
Inspired by the effectiveness of graph self-supervised learning tasks~\cite{jin2021automated}, we propose measuring the competence of expert models on specific graph datasets using the models' self-supervised learning loss values. Specifically, for an input graph $\graph = (\setv, \sete)$, the routing mechanism $\psi$ calculates the dot-product-based relatedness scores for some positive edges $(v_{c_1}, v_{p_1}), \cdots, (v_{c_S}, v_{p_S}) \in \sete$ and analogously calculates the relatedness scores for some sampled negative node pairs $(v_{c_1}, v_{n_1}), \cdots, (v_{c_S}, v_{n_S}) \notin \sete$. The following score difference is then calculated as the competence indicator $\varphi_k$ for the $k$-th expert model regarding the input graph $\graph$:
\begin{align}
    \varphi_{k} = \frac{1}{S}\cdot\sum_{s=1}^S \sigma(\hat{\vece}_{c_s}^\top \hat{\vece}_{p_s} - \hat{\vece}_{c_s}^\top \hat{\vece}_{n_s})
\end{align}
where $\sigma(\cdot)$ represents the sigmoid activation function, which constrains the competence score to the range of (0, 1). This prevents the few outlier cases where the non-activated score difference is excessively large or small, which could otherwise distort the results.\\\vspace{-0.13in}

\noindent \textbf{Training Frequency Regularization}.
Although being empirically accurate in measuring models' competence using the aforementioned self-supervised task loss, this routing mechanism tends to result in a winner-takes-all sub-optimal situation. In this scenario, a single model, or very few models, is predominantly selected as the most competent expert and is used to handle almost all input graphs. These expert models generally receive more or better training samples in the early training stages, giving them an advantage over other experts. Consequently, subsequent training samples are also mostly assigned to them due to their performance advantages, ultimately causing other experts to remain largely untrained.

This situation contradicts our motivation of using different expert models to learn different subsets of graph modeling knowledge. To address this, we propose a training frequency regularization approach that recalibrates the competence score as follows:
\begin{align}
    \varphi'_k = \varphi_k \cdot \left((1 - \frac{m_k}{\sum_{k'}m_{k'}}) \cdot \rho + 1.0 - \frac{\rho}{2} \right)
\end{align}
where $\varphi'k$ represents the recalibrated routing score for the $k$-th expert model $f{\param_k}$, based on the number of previously assigned training steps $m_k$ for $k = 1, \cdots, K$. The notation $\rho$ refers to a hyperparameter for the recalibration scale. A larger $\rho$ results in a greater adjustment to the competence score $\varphi_k$. With this additional step, the expert routing mechanism will assign more training instances to the less trained expert models, thereby preventing the aforementioned winner-takes-all situation.

\subsubsection{\bf Fast Adaptation Capabilities of \model}
With the aforementioned MoE architecture and routing mechanism, the training and inference process of \model\ is conducted by only one expert model. This approach consumes only $1/K$ of the computational and memory resources required for predictions and optimization, compared to other non-MoE graph foundation models based on complex networks like transformers. This endows \model\ with the advantage of fast adaptation when dealing with new datasets.

\subsection{Adaptive and Efficient Graph Experts}
% Unified Structure and Feature Representation
% Simple MLP-based Prediction
% efficient, universal
\subsubsection{\bf Addressing In-domain Graph Heterogeneity}
To handle graph data with different adjacency and feature dimensionalities,  the expert models of our \model\ employ a structure and feature unification process. Adjacency matrices and node features of varying sizes are both mapped into initial node embeddings of fixed dimensionality using a unified mapping function. Inspired by the effectiveness of singular value decomposition (SVD) in extracting important latent features, we utilize SVD for this unified mapping process as follows:
\begin{align}
    &\matu_\mata, \Lambda_\mata, \matv_\mata = \text{SVD}(\tilde{\mata}) ~~~~~~ \matu_\matf, \Lambda_\matf, \matv_\matf = \text{SVD}(\matf)\nonumber\\
    \mate_0 = &\text{LayerNorm}\left(\matu_\mata \sqrt{\Lambda_\mata} + \matv_\mata \sqrt{\Lambda_\mata} + \text{Flip}(\matu_\matf \sqrt{\Lambda_\matf})\right)
\end{align}
% \noindent\textbf{SVD features}. 
Here, $\matu_\mata, \matu_\mata\in\dmnr^{|\setv|\times d}$ and $\matu_\matf\in\dmnr^{|\setv|\times d}, \matv_\matf\in\dmnr^{d_0\times d}$ refer to the $d$-dimensional features obtained through SVD of the Laplacian-normalized adjacency matrix $\tilde{\mata}$~\cite{kipf2016semi} and the node feature matrix $\matf$, respectively. If the dimensionality of $\tilde{\mata}$ or $\matf$ is less than $d$, SVD uses a smaller rank $d'$ equal to the smallest dimensionality of $\tilde{\mata}/\matf$, and the remaining dimensions are padded with zeros up to $d$.\\\vspace{-0.13in}

% \noindent\textbf{Stable patterns of SVD features}. 
Due to the nature of SVD, the dimensions of these features ($\matu_*, \matv_*$) are ranked from the most important to the least important, corresponding to the descending eigenvalues in the diagonal matrices $\Lambda_\mata$ and $\Lambda_\matf$. 
In light of this characteristic, we propose to better preserve the most important feature dimensions for both $\tilde{\mata}$ and $\matf$. In particular, the function $\text{Flip}(\cdot)$ reverses the $d$ dimensions of each row for the SVD features of $\matf$, such that the important features of $\tilde{\mata}$ are aligned with the less important features of $\matf$, and vice versa.\\\vspace{-0.12in}

\noindent\textbf{High-order Connectivity Injection}. 
A non-trainable layer normalization $\text{LayerNorm}(\cdot)$ is applied for numerical stability. The initialized embeddings, denoted as $\mate_0\in\dmnr^{|\setv|\times d}$, have consistent representation dimensionality and relatively stable semantics across datasets.
% \noindent\textbf{High-order connectivity injection}. 
To better preserve the multi-hop connection information into the initial embeddings, \model\ adopts a simplified GCN without parameters~\cite{wu2019simplifying} for $\mate_0$ as follows:
\begin{align}
    \mate_1 = \sum_{l=1}^L \mate_0^{(l)},~~ \mate_0^{(l)}=\tilde{\mata}\cdot\mate_0^{(l-1)},~~\mate_0^{(0)}=\mate_0
\end{align}

% \subsubsection{\bf Simple MLP-based Feature Extraction}
\subsubsection{\bf Efficient and Strong Feature Encoder}
To achieve efficiency while retaining the capacity to encode graph features, our graph experts are configured by deep multi-layer perceptron (MLP) networks. Specifically, the final node embeddings given by an expert model is calculated iteratively as follows:
\begin{align}
    \bar{\mate}^{(l+1)}=\text{LayerNorm}\left(\text{Dropout}\left(\text{ReLU}(\bar{\mate}^{(l)}\matw + \vecb)\right) + \bar{\mate}^{(l)}\right)
\end{align}
The final embeddings are denoted as $\hat{\mate}=\bar{\mate}^{(L')}\in\dmnr^{|\setv|\times d}$, where $L'$ represents the number of fully-connected layers. And $\bar{\mate}^{(0)}$ is initialized by the aforementioned embeddings $\mate_1$. Each layer of our MLP module comprises a linear transformation $\matw\in\dmnr^{d\times d}$ and bias $\vecb\in\dmnr^d$, followed by a ReLU non-linear activation, a dropout layer, a residual connection, and layer normalization.\\\vspace{-0.12in}

\noindent\textbf{Multiple Simple Experts as Strong Encoder}.
It is worth noting that each graph expert in \model\ adopts a very simple learnable network, foregoing the capacity to mine complex hidden relations like those in heavy graph neural networks such as GATs~\cite{velivckovic2017graph} and GraphTransformers~\cite{hu2020heterogeneous}. This is because \model\ employs a MoE architecture, where each expert is expected to handle only a sub-domain of all graph data through simple feature transformations. Therefore, no complex models are needed to accommodate different types of graphs within a single network. Compared to other graph foundation models that rely on a single heavy network, this approach further accelerates the training and inference processes.

\subsection{Efficient Cross-domain Model Training}
% Loss function, cross-graph training algorithm, preprocessing for feature
To maximize the cross-graph generalizability of \model, the training samples from different datasets are mixed together and randomly shuffled for model training. Each batch of training samples contains the following information:
\begin{align}
    \sets=\left(~~~
    \begin{aligned}
        &\{ (v_{c_b},v_{p_b})|b\in B \} \subset \sete_{\graph_s},\\
        &\mate_1 = \text{InitialEmbed}(\graph_s), \\
        &f_{\param_k} ~~\text{where}~~ k=\psi(\graph_s) 
    \end{aligned}
    ~~~\right)
\end{align}
Inspired by the effectiveness of link-wise graph pre-training tasks~\cite{jin2021automated}, we utilize link prediction as the training task. Here, $(v_{c_b}, v_{p_b})$ denotes the positive edges for link prediction, and $B$ denotes the batch size. To facilitate batch training, each training batch involves only one training graph $\graph_s$. The initial node embeddings $\mate_1$ and the most competent expert model $f_{\param_k}$ are preprocessed in advance to accelerate the training. Specifically, the loss function used by our \model\ training is as follows:
\begin{align}
    \loss = \sum_\sets \sum_{b\in B} -\frac{1}{B} \log \frac{\exp (\hat{y}_{c_b, p_b} - \hat{y}_\text{max})}{\sum_{v_{n}\in\setv_{\graph_s}} \exp (\hat{y}_{c_b,n} - \hat{y}_\text{max}) }
\end{align}
This training objective maximizes the prediction scores for positive samples $(v_{c_b}, v_{p_b})$ and minimizes the predictions for all possible node pairs between $v_{c_b}$ and all nodes $v_n$. To avoid numerical instability, we employ a technique where the maximum prediction score of the batch, $\hat{y}_\text{max}$, is subtracted from all prediction scores. 
% The detailed training process is elaborated in Algorithm~\ref{alg:training}.

\subsubsection{\bf Feature and Structure Augmentation}
% Training with Re-projection and Re-assignment
% Data-size-adaptive augmentation frequency
To further enrich the training data, the training of \model\ undergoes periodic reprocessing of, \textbf{firstly}, the initial graph embeddings $\mate_1$, and \textbf{secondly}, the graph routing results. We demonstrate that such reprocessing augments the features and structures of the original graph data, thereby training \model\ using more diversified input data.

For the initial graph embeddings, we periodically reconduct the SVD and simplified GCN processes after a certain number of training steps. This helps generate different embedding spaces for the same data, thereby greatly improving the generalizability of \model\ regarding representation heterogeneity~\cite{li2024zerog}. To prevent this process from consuming excessive computational time, we propose adopting different augmentation frequencies adaptive to the size of different datasets. Specifically, each dataset undergoes this representation augmentation after $|\sete|/(10B)$ training steps.

For the graph routing results, we also periodically recalculate the recalibrated competence scores. Specifically, the positive sample pairs $(v_{c_s}, v_{p_s})$ for $s=1, \cdots, S$, as well as the negative samples $v_{n_s}$, are randomly sampled. This essentially performs structure augmentation by using a random subset to evaluate the performance of graph experts on the input graph, thereby enhancing the model's robustness against structural noise.

\subsubsection{\bf Complexity Analysis}
The training and inference process of \model\ is conducted by only one expert model, which has a complexity of $\mathcal{O}(B \times d^2 \times L')$ for each batch. Since we preprocess the initial embeddings and the expert routing, these two processes do not increase the batch-wise computational complexity. As a result, the complexity of the forward and backward steps for \model\ is much lower than that of other graph foundation models that involve complex GNNs and graph transformers.
Additionally, the expert routing performs $\mathcal{O}\left(\sum_{\graph_s} |\sete_s| \times d \times K + \sum_{\graph_s} |\setv_s| \times d^2 \times L' \times K\right)$ computations, where the latter term empirically has a larger scale compared to the former term. This dominant term is similar to a simple GCN network of a comparable model size.
Overall, \model\ is more efficient than existing methods in both training and inference, and the additional computations for routing have a complexity comparable to simple GNNs.
% One major efficiency advantage of \model\ is that our method needs only to consume $1/K$ computational and memory resources to make predictions, compared to other non-MoE graph foundation models based on complex networks like transformers.
\section{Evaluation}
\label{sec:eval}
Our experiments aim to answer the following \textbf{R}esearch \textbf{Q}uestions:
\begin{itemize}[leftmargin=*]
    \item \textbf{RQ1}: How does the zero-shot prediction performance of \model\ compare to different baseline methods?%\\\vspace{-0.12in}
    \item \textbf{RQ2}: How do \model's various modules impact its overall performance with the contribution of each component?%\\\vspace{-0.12in}
    \item \textbf{RQ3}: How does the model size and the amount of training data influence the performance of \model?%\\\vspace{-0.12in}
    \item \textbf{RQ4}: How interpretable is the expert routing mechanism within \model's graph Mixture-of-Experts (MoE) architecture?%\\\vspace{-0.12in}
    % \item \textbf{RQ5}: On which types of data characteristics does the MoE architecture of our \model\ bring bigger benefits?
    \item \textbf{RQ5}: How is the scalability and efficiency of \model\ compare to fine-tuning methods when adapting to new datasets?
\end{itemize}

\subsection{Experimental Settings}
\subsubsection{\bf Experimental Datasets}
To conduct a comprehensive evaluation of the cross-domain generalizability of graph models, we employ a total of \textbf{38} graph datasets. These datasets span a wide range of domains, including e-commerce (\eg~user interactions and product-wise relations), academic graphs (\eg~citation and collaboration networks), biological information networks (\eg~relations among drugs and proteins), and other domains like email networks, website networks, trust networks, and road networks.

We set up different dataset groups and conduct cross-dataset evaluations on these groups. Specifically, all datasets are divided into two cross-domain groups, \textbf{Link1} and \textbf{Link2}, which have a similar number of total edges and a similar number of domain-specific edges. 
Specifically, the Link1 and Link2 groups contain 15 and 18 datasets, respectively.
For the node classification task, we use 5 datasets gathered from e-commerce and academic information scenarios.
Additionally, we have three domain-specific groups: \textbf{Ecommerce}, \textbf{Academic}, and \textbf{Others}. The \textbf{Others} group is primarily composed of biological networks, combined with other small domains that have fewer datasets. 
% Node and graph datasets
See Appendix~\ref{app:datasets} for more information of our experimental datasets.

\subsubsection{\bf Experimental Settings}
We follow previous works~\cite{he2020lightgcn,kipf2016semi} for dataset splitting and evaluation metrics. Our \model\ model and the graph foundation models are evaluated on a cross-graph zero-shot prediction task. For baselines that cannot handle cross-dataset transfer, we evaluate their few-shot performance. Details of the evaluation protocols are provided in Appendix~\ref{app:protocols}. The \textbf{Hyperparameter Settings} of \model\ are provided in Appendix~\ref{app:hyperparam}. The compared \textbf{Baseline Methods} are introduced in Appendix~\ref{app:baselines}.

% \subsubsection{\bf Evaluation Protocols} 
% We follow previous works~\cite{} for dataset splitting and evaluation metrics. Our \model\ and the graph foundation models are evaluated on a cross-graph zero-shot prediction task. For baselines that cannot handle cross-dataset transfer, we evaluate their few-shot performance. Details of the evaluation protocols are provided in Appendix~\ref{app:protocols}.

% \subsubsection{\bf Hyperparameter Settings \& Baselines}
% Detailed configurations of our \model\ are provided in Appendix~\ref{app:hyperparam}. And the compared baselines are introduced in Appendix~\ref{app:baselines}.

\begin{table*}[t]
    \centering
    \caption{We evaluate the \model\ model (in zero-shot settings) and baseline models (with 5\% and 10\% training data) on link prediction (Recall@20, NDCG@20), node classification (Accuracy, Macro F1), and graph classification (Accuracy, Macro F1).}
    \label{tab:overall_performance}
    \small
    \setlength{\tabcolsep}{0.5mm}
    \vspace{-0.1in}
    \begin{tabular}{c|cc|cc|cc|cc|cc|cc|cc|cc|cc|cc|cc}
        \hline
        \multirow{2}{*}{Data} & \multicolumn{4}{c|}{GIN} & \multicolumn{4}{c|}{GAT} & \multicolumn{4}{c|}{GPF} & \multicolumn{4}{c|}{GraphPrompt} & \multicolumn{4}{c|}{GraphCL} & \multicolumn{2}{c}{\textbf{\model}}\\
        \cline{2-23}
        & \multicolumn{2}{c|}{Train 5\%} & \multicolumn{2}{c|}{Train 10\%} & \multicolumn{2}{c|}{Train 5\%} & \multicolumn{2}{c|}{Train 10\%} & \multicolumn{2}{c|}{Tune 5\%} & \multicolumn{2}{c|}{Tune 10\%} & \multicolumn{2}{c|}{Tune 5\%} & \multicolumn{2}{c|}{Tune 10\%} & \multicolumn{2}{c|}{Tune 5\%} & \multicolumn{2}{c|}{Tune 10\%} & \multicolumn{2}{c}{0-shot}\\
        \hline
        \hline
        {Metric} & Rec & NDCG & Rec & NDCG & Rec & NDCG & Rec & NDCG & Rec & NDCG & Rec & NDCG & Rec & NDCG & Rec & NDCG & Rec & NDCG & Rec & NDCG & Rec & NDCG\\
        \hline
        {Link1} & 6.46 & 3.06 & 11.80 & 5.45 & 13.52 & 6.65 & 13.45 & 6.78 & 6.04 & 2.92 & 6.80 & 3.27 & 4.33 & 2.24 & 5.42 & 3.11 & 17.23 & 9.00 & 20.55 & 10.76 & \textbf{23.94} & \textbf{12.68}\\
        \hline
        {Link2} & 6.72 & 4.50 & 21.62 & 13.41 & 9.83 & 5.91 & 15.30 & 8.84 & 7.44 & 4.25 & 16.58 & 9.84 & 6.06 & 3.36 & 6.10 & 3.62 & 29.18 & 17.62 & 31.42 & 19.91 & \textbf{46.42} & \textbf{27.21}\\
        \hline
        {Ecom.} & 3.36 & 2.58 & 13.41 & 8.06 & 3.79 & 2.94 & 9.64 & 5.78 & 7.25 & 3.84 & 18.72 & 10.94& 4.90 & 2.59 & 6.06 & 3.36 & 22.13 & 13.19 & 26.05 & 14.59 & \textbf{26.92} & \textbf{15.05}\\
        \hline
        {Acad.} & 10.82 & 4.70 & 20.61 & 9.04 & 14.95 & 6.29 & 11.17 & 4.67 & 13.22 & 5.80 & 14.83 & 6.41 & 6.73 & 3.05 & 7.72 & 3.40 & 24.86 & 12.50 & 28.69 & 14.31 & \textbf{32.74} & \textbf{15.31}\\
        \hline
        {Othrs.} & 6.92 & 4.46 & 18.43 & 11.85 & 16.34 & 9.22 & 16.17 & 20.88 & 2.40 & 2.12 & 4.51 & 3.44 & 2.93 & 2.36 & 3.42 & 2.72 & 24.54 & 14.93 & 24.62 & 15.90 & \textbf{46.83} & \textbf{28.97}\\
        \hline
        \hline
        {Metric} & Acc & MacF1 & Acc & MacF1 & Acc & MacF1 & Acc & MacF1 & Acc & MacF1 & Acc & MacF1 & Acc & MacF1 & Acc & MacF1 & Acc & MacF1 & Acc & MacF1 & Acc & MacF1\\
        \hline
        {Node} & 20.79 & 19.46 & 36.04 & 30.60 & 53.76 & 40.14 & 54.83 & 41.61 & 12.77 & 11.45 & 16.29 & 16.00 & 18.01 & 20.59 & 23.15 & 22.89 & 43.70 & 33.72 & 48.75 & 36.15 & \textbf{64.31} & \textbf{43.24}\\
        \hline
        % {Graph} &  &\\
        % \hline
    \end{tabular}
    \vspace{-0.1in}
\end{table*}

\begin{table}[]
    \centering
    \small
    \caption{Comparing \model\ framework to existing graph foundation models in zero-shot prediction capabilities.}
    \label{tab:foundation_models}
    \setlength{\tabcolsep}{1.6mm}
    \vspace{-0.1in}
    \begin{tabular}{l|c|c|c|c|c|c}
        \hline
        Method & \multicolumn{4}{c|}{GraphGPT} & \multicolumn{2}{c}{OpenGraph} \\
        \hline
        Data & \multicolumn{2}{c|}{Pubmed} & \multicolumn{2}{c|}{Cora} & \multicolumn{2}{c}{Ecom. w/o GR}\\
        % Data & \multicolumn{2}{c}{\underline{Pubmed + Cora}} & \multicolumn{2}{c}{\underline{Ecom. - Goodreads}}\\
        \hline
        Metric & Acc & MacF1 & Acc & MacF1 & Recall & NDCG \\
        % \midrule
        \hline\hline
        Baseline & 0.1813 & 0.1272 & 0.7011 & 0.6491 & 0.1444 & 0.1099\\
        \model-F & 0.5852 & 0.5325 & 0.7134 & 0.6003 & 0.2281 & \textbf{0.1600}\\
        \model & \textbf{0.6088} & \textbf{0.5492} & \textbf{0.7809} & \textbf{0.7591} & \textbf{0.2382} & 0.1552\\
        \hline
    \end{tabular}
    \vspace{-0.1in}
\end{table}

\subsection{\model's Zero-Shot Prediction (RQ1)}
% We assess the zero-shot prediction capabilities of our model across 38 graph datasets from various domains. Two versions of \model\ are independently trained—one on the Link1 dataset and the other on Link2. Each model is subsequently used to make zero-shot predictions on the dataset it was not originally trained with. It is important to note that Link1 and Link2 datasets do not share the same feature spaces or sources of data collection. The outcomes of this evaluation are detailed in Table~\ref{tab:overall_performance} and Table~\ref{tab:foundation_models}. Our principal observations are summarized as follows:
To assess the zero-shot prediction capabilities of the \model\ model, we conducted an extensive evaluation across 38 graph datasets from various domains. We independently trained two versions of the \model\ model - one on the Link1 dataset and the other on the Link2 dataset. Each trained model was then used to make zero-shot predictions on datasets it was not originally trained with. It is important to note that the Link1 and Link2 datasets do not share the same feature spaces or sources of data collection, which adds to the complexity and challenges of the zero-shot evaluation. We also compare our \model\ with existing graph foundation models. And in this comparison we add another \model-F version, which removes the utilization of node features. The outcomes of this evaluation are detailed in Table~\ref{tab:overall_performance} and Table~\ref{tab:foundation_models}, and our key observations are listed as follows: \\\vspace{-0.12in}

\noindent \textbf{i) Superior Generalizability across Diverse Datasets.} $\bullet$ \textbf{Superior Prediction Accuracy}. Compared to the few-shot capabilities of existing GNN models, pre-training techniques, and foundation models, \model\ demonstrates exceptional zero-shot prediction accuracy across various domains. This superior performance spans both link prediction and node classification tasks. $\bullet$ \textbf{Effectively Handling Heterogeneity}. The enhanced generalizability can be attributed to the effective handling of structure-level and feature-level data heterogeneity through unified structure and feature representations in the expert models. This approach enables \model\ to develop comprehensive modeling functions that are universally applicable across different graph data scenarios. $\bullet$ \textbf{Comprehensive Training}. Additionally, the extensive training regimen, which incorporates a variety of large-scale datasets, equips \model\ with a deep and broad expertise in graph modeling and prediction.\\\vspace{-0.12in}

\noindent \textbf{ii) Limitation of existing pre-training GNNs.}
$\bullet$ \textbf{Challenges of Cross-Domain Transfer}. Existing pre-training and tuning methods, like GPF, GraphPrompt, and GraphCL, employ self-supervised learning and are pre-trained on half the datasets, then fine-tuned on the remaining datasets using few-shot data. However, this pre-training often fails to yield significant improvements due to substantial distribution disparities across data domains. For instance, datasets may exhibit vastly different link densities or utilize distinct node features, which significantly challenges the transfer of useful knowledge from divergent pre-training datasets during fine-tuning and prediction. $\bullet$ {\model's Robust Adaptability} To address this challenge, the AnyGraph model incorporates multiple graph expert models tailored to various sub-domains of graph data. This MoE architecture effectively manages datasets from distinctly different domains, such as e-commerce user behaviors, academic networks, and road networks, demonstrating its robust adaptability.\\\vspace{-0.12in}
% Existing pre-training and tuning methods, such as GPF, GraphPrompt, and GraphCL, employ self-supervised learning objectives and are initially pre-trained on half of all datasets (\ie, Link1 and Link2), followed by tuning on the remaining datasets using few-shot data. However, this pre-training often fails to yield significant improvements, primarily due to the substantial distribution disparities across different data domains. For instance, two datasets might exhibit vastly different link densities or utilize entirely distinct node features. These cross-domain discrepancies significantly challenge the transfer of useful knowledge from highly divergent pre-training datasets during subsequent fine-tuning and prediction phases. To address this challenge, \model\ incorporates multiple graph expert models tailored to various sub-domains of graph data. This MoE architecture effectively manages datasets from distinctly different domains, such as e-commerce user behaviors, academic networks, and road networks, demonstrating its robust adaptability.\\\vspace{-0.12in}

% \noindent \textbf{iii) Difficulty of handling multi-domain graph data}

\subsection{Scaling Law of \model\ Framework (RQ2)}
\begin{figure*}
    \centering
    \subfigure[Cross-domain performance on Link1 (zero-shot) and Link2 (full-shot).]{
        \includegraphics[width=0.16\textwidth]{figs/param_zeroshot_multidomain.pdf}
        \includegraphics[width=0.16\textwidth]{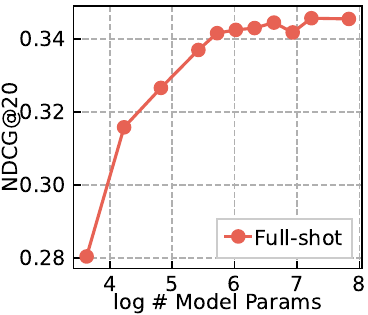}
        \includegraphics[width=0.16\textwidth]{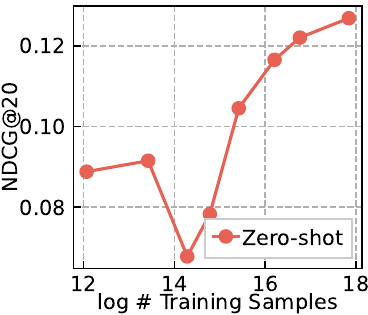}
    }
    \subfigure[Performance on academic data.]{
        \includegraphics[width=0.16\textwidth]{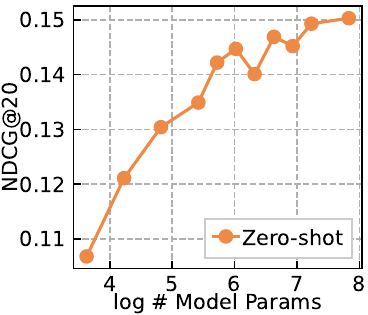}
        \includegraphics[width=0.16\textwidth]{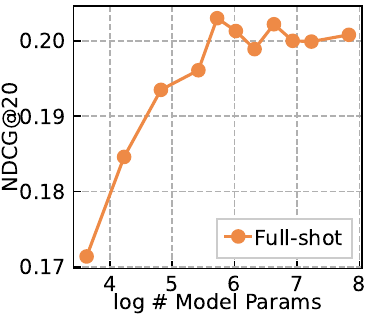}
        \includegraphics[width=0.16\textwidth]{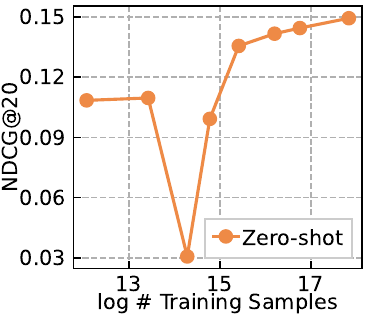}
    }
    \subfigure[Performance on ecommerce data.]{
        \includegraphics[width=0.16\textwidth]{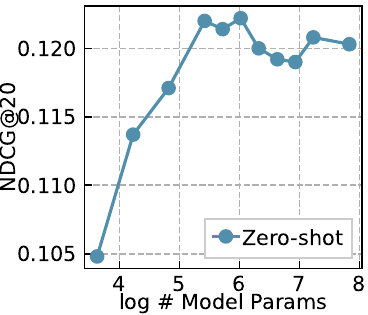}
        \includegraphics[width=0.16\textwidth]{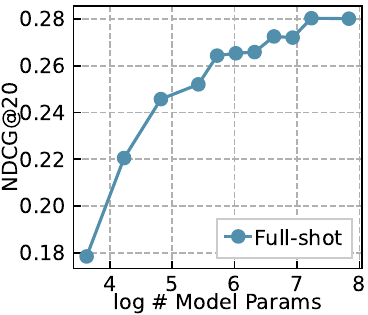}
        \includegraphics[width=0.16\textwidth]{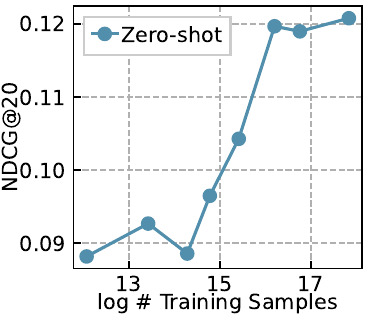}
    }
    \subfigure[Performance on others.]{
        \includegraphics[width=0.16\textwidth]{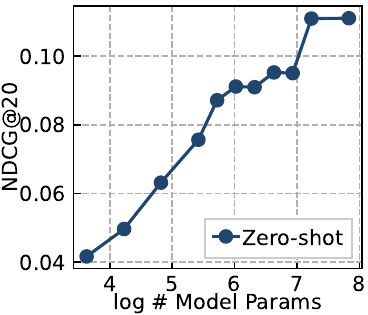}
        \includegraphics[width=0.16\textwidth]{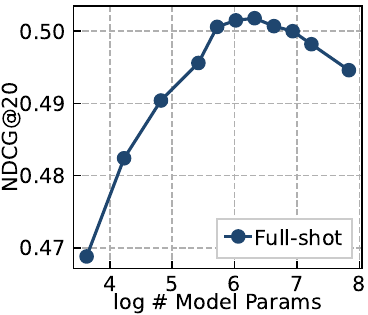}
        \includegraphics[width=0.16\textwidth]{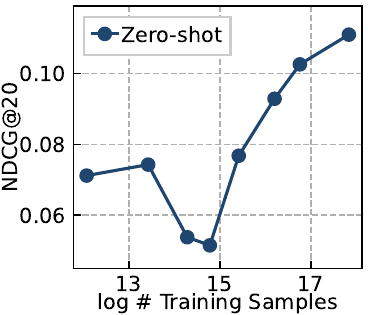}
    }
    \vspace{-0.15in}
    \caption{Zero-shot and full-shot performance \textit{w.r.t.} the number of model parameters and the amount of training samples.}
    \label{fig:scaling_law}
    % \vspace{-0.05in}
\end{figure*}

In this section, we explore the applicability of the scaling law to \model. We conduct experiments using 18 different versions of \model, each differing in model size and quantity of training data. Specific configurations of these variants are discussed in Appendix~\ref{app:scaling_law}. The evaluation results are depicted in Figure~\ref{fig:scaling_law}, which includes overall and domain-specific performance, as well as zero-shot and full-shot outcomes. Our key findings are as follows:\\\vspace{-0.12in}

\noindent\textbf{i) Generalizability of \model\ Follows the Scaling Law}. As the model size and the volume of training data increase, we notice a saturation point in \model's full-shot performance. In contrast, the zero-shot prediction accuracy continues to improve. This pattern supports the scaling law of graph foundation models, illustrating that scaling up can significantly enhance the capabilities of graph models. Two key factors contribute to this phenomenon:
\begin{itemize}[leftmargin=*]
    \item \textbf{Task Difficulty}. The saturation in full-shot performance is partly because the evaluation tasks might not be challenging enough. In-domain generalization can be more straightforward, leading to a plateau in performance improvements. This insight into the scaling law for graph data encourages further exploration of larger models on more complex graph learning tasks.
    \item \textbf{MoE Architecture}. The integration of the Mixture of Experts (MoE) architecture allows \model\ to effectively manage and utilize a broader spectrum of knowledge, particularly in this zero-shot scenario characterized by significant distribution disparities.
\end{itemize}

\noindent\textbf{ii) Emergent Abilities of \model}. The overall zero-shot performance curve illustrates that as the model size increases, the performance sometimes experiences periodic stagnation. With further increments in parameters, \model's performance undergoes a sudden significant improvement. This phenomenon indicates the emergent abilities of \model, demonstrating the effectiveness of scaling up in enhancing its generalization capabilities.\\\vspace{-0.12in}

\noindent\textbf{iii) Insufficient training data may bring bias}. In the initial stages of increasing the training data, the introduction of new datasets might negatively impact performance due to their differences from the test graphs. However, this issue can be mitigated by further expanding the training data. By providing the model with a more comprehensive set of training samples, it helps prevent overfitting and reduces bias stemming from dataset disparities.

\begin{figure}
    \centering
    \includegraphics[width=0.9\columnwidth]{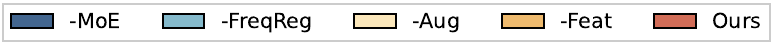}
    \vspace{-0.03in}
    
    \hspace{-0.05in}
    \subfigure[Multi-domain zero-shot performance.]{
        \includegraphics[width=0.24\columnwidth]{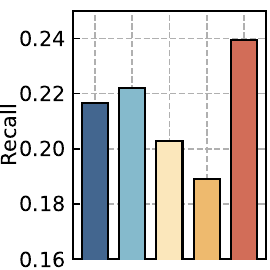}
        \includegraphics[width=0.24\columnwidth]{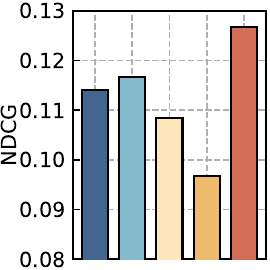}
    }
    \hspace{-0.1in}
    \subfigure[Multi-domain full-shot performance.]{
        \includegraphics[width=0.24\columnwidth]{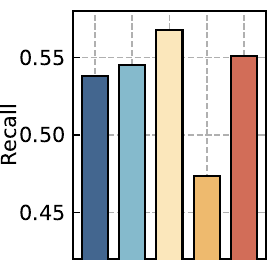}
        \includegraphics[width=0.24\columnwidth]{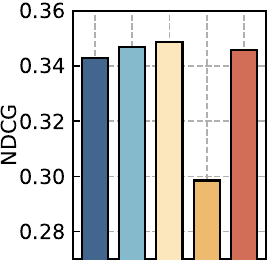}
    }

    \hspace{-0.05in}
    \subfigure[Zero-shot performance on Academic.]{
        \includegraphics[width=0.24\columnwidth]{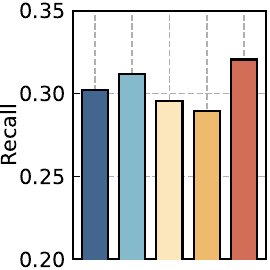}
        \includegraphics[width=0.24\columnwidth]{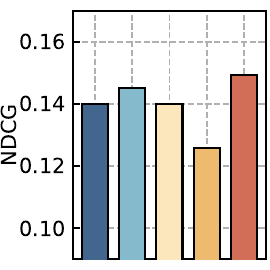}
    }
    \hspace{-0.1in}
    \subfigure[Full-shot performance on Academic.]{
        \includegraphics[width=0.24\columnwidth]{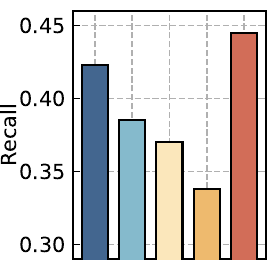}
        \includegraphics[width=0.24\columnwidth]{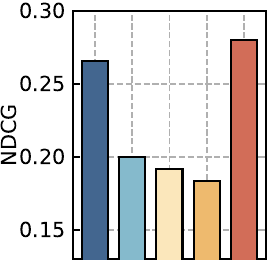}
    }
    \vspace{-0.2in}
    \caption{Impact of different sub-modules on the zero-shot and full-shot prediction capabilities of \model.}
    \label{fig:ablation}
    \vspace{-0.2in}
\end{figure}

\subsection{Ablation Study (RQ3)}
This section evaluates the effectiveness of \model's sub-modules by comparing ablated variants in terms of their zero-shot and full-shot performance across both cross-domain datasets and domain-specific datasets (specifically Academic data). The results are in Figure~\ref{fig:ablation}. We make the following observations:

\begin{itemize}[leftmargin=*]
    \item \textbf{MoE Significantly Enhances Zero-Shot Performance}. The \textbf{-MoE} variant, which employs a single expert model without the MoE architecture, demonstrates  decent performance on datasets on which it was trained, as shown in parts (b) and (c). However, this variant exhibits a substantial decline in zero-shot prediction capabilities. This underscores the critical role of the MoE architecture in enhancing \model's generalization abilities. The use of multiple expert models significantly expands \model's modeling capacity, effectively managing the large disparities between various domains using multiple seperated models.\\\vspace{-0.12in}
    \item \textbf{Feature Modeling is Crucial in \model}. In the -Feat variant, node features are omitted, leading to the most significant degradation in both zero-shot and full-shot performance. This underscores the effectiveness of \model’s unified structure and feature representation method in successfully learning features. This component is crucial for tackling in-domain graph data heterogeneity. Additionally, this outcome highlights the feasibility of unifying different feature spaces created by various methods into a single model for general use.\\\vspace{-0.12in}
    \item \textbf{Effectiveness of Frequency Regularization and Graph Augmentation}. In the \textbf{-FreqReg} and \textbf{-Aug} variants of \model, the routing adjustment based on the training frequency of experts and the feature and structure augmentation are individually removed. 
    Frequency regularization specifically prevents suboptimal training outcomes by ensuring that all experts are utilized, avoiding scenarios where the majority are overlooked. Meanwhile, the graph augmentation module plays a crucial role in enriching the training graph data, thus enhancing the model's robustness.
    The outcomes from these modifications affirm the beneficial impact of these two components within \model. Omitting them can lead to biased model training, which undermines the robustness of \model\ in handling diverse datasets.
\end{itemize}

\subsection{Investigation on Expert Routing (RQ4)}
\begin{figure}
    \centering
    % \subfigure[Full-shot datasets.]{
        \includegraphics[width=0.45\columnwidth]{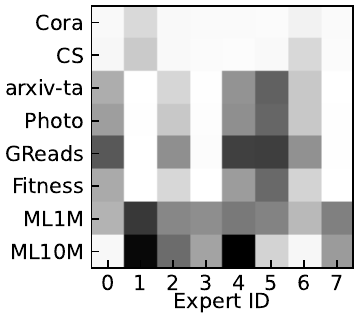}
    % }
    % \subfigure[Zero-shot datasets.]{
        \includegraphics[width=0.45\columnwidth]{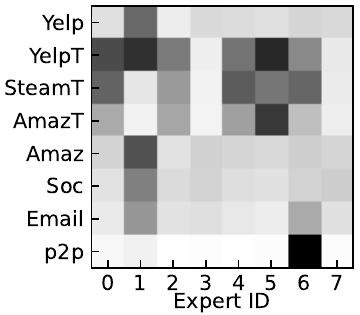}
    % }
    \vspace{-0.15in}
    \caption{Competence score between datasets and expert models, given by the routing mechanism of \model.}
    \label{fig:case_study}
    \vspace{-0.15in}
\end{figure}

This section delves into the expert routing mechanism of \model. Figure~\ref{fig:case_study} displays the competence scores of various expert models for the input datasets, as determined by \model’s routing algorithm based on the self-supervised loss. The figure illustrates that datasets sharing common characteristics—such as source of collection or feature construction method—are often routed to the same expert models by \model. For instance, datasets like arxiv-ta, Photo, GoodReads, and Fitness, which utilize a common text-embedding-based feature space, are assigned to highly similar experts
(expert 0, 2, 4, 5). 
Additionally, ML1M and ML-10M, both sourced from the movie-rating platform Movielens, are predominantly associated with expert 1. It is also notable that this routing pattern extends to zero-shot datasets, as shown on the right part of Figure~\ref{fig:case_study}. Here, YelpT, SteamT, and AmazonT, which share the same feature space, are assigned to very similar expert models. 
This outcome underscores the efficacy of \model's routing mechanism in identifying the appropriate expert models for various datasets, and also showcases its explainability in revealing graph-wise relatedness.
% Additionally, this expert assignment enhances the model's explainability by clarifying the relatedness of the input graph data.
% This outcome highlights the effectiveness and the explainability of \model's routing mechanism.

\subsection{Efficiency Study (RQ5)}
\begin{figure}
    \centering
    \subfigure[Performance on Citation-2019.]{
        \includegraphics[width=0.47\columnwidth]{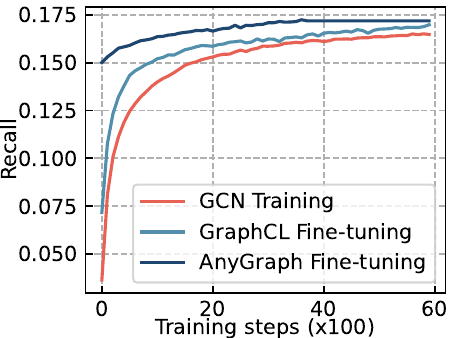}
    }
    \subfigure[Performance on PPA.]{
        \includegraphics[width=0.47\columnwidth]{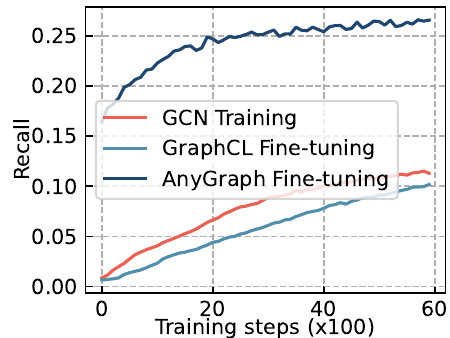}
        \label{fig:finetune_ppa}
    }
    \vspace{-0.15in}
    \caption{Performance v.s. training/tuning steps.}
    \label{fig:finetune}
    \vspace{-0.1in}
\end{figure}

\noindent\textbf{Tuning Curve Comparison}. To evaluate the efficiency of \model, we compare its fine-tuning process with that of GraphCL and the training from scratch process of a GCN model. As depicted in Figure~\ref{fig:finetune}, when fine-tuned on a new dataset, the pre-trained \model\ rapidly achieves a high performance saturation point. In some instances, such as with the PPA dataset, GraphCL and the end-to-end trained GCN struggle to attain comparable performance levels. This advantage is based on i) the strong cross-domain generalization capabilities of \model, which bring a high starting point for the new dataset, and ii) the efficiency of \model's MoE architecture, which requires only one MLP network for efficient but effective modeling and parameter tuning.

In addition, it is observed that pre-training GraphCL does not consistently benefit its fine-tuning on new datasets, as evidenced by GraphCL’s underperformance relative to GCN in Figure~\ref{fig:finetune_ppa}. This should be ascribed to the large distribution gap between the pre-training data Link2 and the test data PPA. \\\vspace{-0.12in}

\noindent\textbf{Training Time Comparison}. To evaluate the efficiency of the models under consideration, we compared the training times of the three models. As indicated in Table~\ref{tab:training_time}, \model, despite having significantly more parameters, has training times that are comparable to, or even less than, the other two models. This underscores the efficiency of our model design, and demonstrates the efficiency of \model\ to adapt to new data through model tuning.

Specifically, \model\ avoids the cumbersome process of full-graph propagation at each training step. Instead, it utilizes structure-aware embeddings derived through a non-trainable pre-processing method. This approach significantly reduces both the time and memory requirements for \model. Furthermore, the MoE architecture equips \model\ with the capability to use only $1/K$ of the computational resources for most prediction and optimization processes, thereby greatly reducing overall computational costs.

\begin{table}[t]
    \centering
    \caption{Training time for each 100 steps on different data.}
    \setlength{\tabcolsep}{1.0mm}
    \vspace{-0.15in}
    \label{tab:training_time}
    \begin{tabular}{c|cccccccc}
        \hline
        Dataset & CS & ML1M & Yelp & Email & Cite19 & roadNet & PPA \\
        \hline
        \hline
        GCN & 1.5s & 4.2s & 6.0s & 2.5s & 19.2s & 27.8s & 101.1s\\
        \hline
        GraphCL & 1.1s & 4.9s & 9.4s & 2.8s & 43.1s & 57.1s & 130.8s \\
        \hline
        Ours & 1.5s & 3.5s & 6.1s & 3.0s & 31.6s & 37.3s & 41.1s\\
        \hline
    \end{tabular}
    \vspace{-0.15in}
\end{table}

\section{Related Works}
\label{sec:relate}

\noindent \textbf{Graph Neural Models}.
Graph learning has garnered significant interest for its broad applicability across various fields such as user behavior modeling, social analysis, and studies in biology and chemistry~\cite{chang2021sequential,hao2020asgn}. Graph neural networks (GNNs) learn node representation vectors for downstream tasks like node classification and link prediction. The core mechanism involves iterative message passing, refining node embeddings to capture both node-specific information and higher-order topological structures. This process ensures that the final node embeddings effectively encapsulate both node-specific information and higher-order topological structures. Notable techniques include Graph Convolutional Networks (GCNs)~\cite{jin2021universal}, Graph Attention Networks (GATs)~\cite{brody2021attentive}, Graph Isomorphism Network (GIN)~\cite{xu2018powerful}, and Graph Transformer~\cite{hu2020heterogeneous}, which improves the encoding function for better graph modeling. Despite these advancements, these methods still require high-quality training data and often struggle with generalization capabilities.\\\vspace{-0.12in}

\noindent \textbf{Self-Supervised Graph Learning}. Given the challenges with the generalizability of GNNs, considerable research efforts~\cite{xie2022self} have focused on enhancing GNNs through self-supervised learning objectives, aiming to capture invariant graph features. Specifically, GraphCL~\cite{you2020graph} introduced a contrastive pre-training approach for graph data, designed to learn authentic graph characteristics that are robust to structural and feature perturbations. Building on this, JOAO~\cite{you2021graph} and GCA~\cite{zhu2021graph} have developed adaptive augmentation strategies for self-supervised tasks, effectively mitigating the adverse effects of random augmentations. 
% Additionally, DGI~\cite{} introduced a mutual information maximization pre-training method that utilizes both local and global graph views for contrastive learning. 
Subsequent works have sought to quickly adapt these pre-trained models to downstream tasks and evolving graph data, as demonstrated by GPF~\cite{fang2022universal} and GraphPrompt~\cite{liu2023graphprompt}. Despite these advancements, the generalizability of these methods remains confined to graph data with similar structural patterns and feature spaces, thus not addressing the cross-domain generalization challenges highlighted in this paper.\\\vspace{-0.12in}

\noindent \textbf{Large-scale Graph Pre-training}.
Recent advances in graph modeling have seen efforts to pre-train large-scale graph models across multiple datasets to improve their generalization abilities, drawing inspiration from the strong generalization capabilities of large language models (LLMs)~\cite{xia2024opengraph}. For instance, OFA~\cite{liuone} and ZeroG~\cite{li2024zerog} utilize text embeddings to standardize the feature spaces across various graph datasets and tasks, facilitating cross-dataset training of graph models. Models like InstructGLM~\cite{ye2024language} GraphGPT~\cite{tang2024graphgpt} and LLaGA~\cite{chen2024llaga} synchronize graph representation spaces with the hidden spaces of LLMs,
% This alignment allows the seamless application of general language model methodologies to graph prediction tasks, opening new avenues for leveraging language processing techniques in graph learning.
thus enabling the application of general LLMs for graph prediction tasks. 
Furthermore, HiGPT~\cite{tang2024higpt} expands the capabilities of LLMs to accommodate heterogeneous graph data.

Despite these advancements, most generalized graph models require substantial access to and integration of text features, which confines their use primarily to text-abundant environments such as academic networks. Additionally, these methods are typically trained within specific application realms, failing to address the significant variances between datasets from diverse domains.
\section{Conclusion}
\label{sec:conclusion}

In this work, the presented \model\ framework, an effective and efficient graph foundation model designed to address the multifaceted challenges of structure and feature heterogeneity across diverse graph datasets. \model's innovative Mixture-of-Experts (MoE) architecture, coupled with its dynamic expert routing mechanism, positions it at the state-of-the-art of cross-domain generalization capabilities. Extensive experiments on 38 varied graph datasets have not only underscored AnyGraph's superior zero-shot learning performance but also its robustness to distribution shifts and its adherence to scaling laws, thereby enhancing its predictive accuracy with increased model size and data volume. The model's efficiency in training and inference, validated through comparison with existing methods, further cements its practical applicability. 

\clearpage

\bibliographystyle{abbrv}
\balance
\bibliography{refs}

\clearpage
\appendix
\section{Appendix}
\label{app}

\subsection{Experimental Datasets}
\label{app:datasets}
We utilize a total of 38 graph datasets across various domains. The entire dataset contains 14,437,372 nodes, and 199,265,688 edges. The dataset specifics are detailed below:

\begin{table*}[]
    \centering
    
    \caption{Statistics of the experimental datasets.}
    \vspace{-0.15in}
    \label{tab:my_label}
    \small
    \setlength{\tabcolsep}{0.2mm}
    \begin{tabular}{c|c|c|c|c|c|c|c|c|c|c|c}
        \hline
        Dataset & DDI & Collab & ML1m & ML10m & Amazon-book & PPA & Yelp2018 & Gowalla & Cora & Pubmed & Citeseer\\
        \hline
        \# Nodes &  4,267& 235,868 & 9,746 & 80,555 & 144,242 & 576,289 & 69,716	& 70,839 & 2,708 & 19,717 & 3,327 \\
        \hline
        \# Edges & 1,334,889 & 1,285,465 & 920,193 & 9,200,050 & 2,984,108 & 45,495,642 & 1,561,406 & 1,027,370 & 10,556& 88,648 &9,104 \\ 
        \hline
        $d$ Feats  & 0 & 128 & 0 & 0 & 0 & 58 & 0 & 0 & 1433& 500 & 3703\\
        \hline
        \hline
        Datasets  & Proteins-0 & Proteins-1 & Proteins-2 & Proteins-3 & Products-home&Products-tech&Yelp-t & Amazon-t & Steam-t & Goodreads & Fitness\\
        \hline
        \# Nodes & 25,449 & 6,568 & 18,108& 13,015 & 9,790 & 47,428& 22,101& 20,332& 28,547& 676,084& 173,055\\
        \hline
        \# Edges & 11,660,646& 1,845,960& 7,418,688 &3,962,930 & 131,843 & 2,077,241 & 277,535& 200,860& 525,922& 8,582,306&1,773,500  \\
        \hline
        $d$ Feats & 0& 0& 0& 0& 100&100 & 1536& 1536& 1536& 768& 768\\
        \hline
        \hline
        Datasets &Soc-Epinions1&Email-Enron&Web-Stanford & RoadNet-PA & P2P-Gnutella06& Citation-2019& Citation-20Century & Arxiv& Arxiv-t& Photo & CS\\
        \hline
        \# Nodes & 75,879 & 36,692 &281,903 &1,088,092& 8,717& 765,658& 1,016,241 & 169,343 & 169343& 48,362 & 18,333\\
        \hline
        \# Edges & 508,837& 183,831&2,312,497&1,541,898& 31,525& 1,917,381& 5,565,798 &1,166,243&1,166,243 &500,939 & 163,788\\
        \hline
        $d$ Fets & 0 &0 & 0& 0& 128& 128 & 128 & 128& 768 & 768 & 6805\\
        \hline
    \end{tabular}
\end{table*}

\noindent \textbf{E-commerce Datasets}. This category includes 15 datasets from various e-commerce contexts such as user rating platforms and online retail services. These datasets vary in terms of the presence and type of node features. For instance, datasets such as {Amazon-book, Yelp2018, Gowalla, Yelp-text, Amazon-text, Steam-text, Goodreads, Amazon-Fitness, Amazon-Photo, Movielens-1M, Movielens-10M, Products-home, Products-tech, Home-node, Tech-node} are included. Notably, Amazon-text, Steam-text, and Yelp-text utilize the same method for feature generation, while Fitness, Photo, and Goodreads employ a different consistent method.

\noindent \textbf{Academic Network Datasets}. We use 13 datasets focused on academic networks, which include citation and collaboration relations among scholars and papers. These datasets represent various research fields and employ diverse feature generation methods, such as NLP embeddings, bag-of-words, and different versions of large language models. The specific datasets are {Cora, Pubmed, Arxiv, Cora-link, Pubmed-link, Citeseer, CS, Arxiv-link, Arxiv-t} (with features derived using an alternative method), {Citation-2019, Citation-20Century, OGB-Collab}.

\noindent \textbf{Biological Information Networks}. Our collection includes 6 datasets related to biological entities like proteins, drugs, and diseases. This category features networks such as {OGB-DDI, OGB-PPA} and four protein relation networks for different species, denoted as {Proteins-0, Proteins-1, Proteins-2, Proteins-3}.

\noindent \textbf{Other Datasets}. In addition to the categories mentioned above, we include 5 datasets from various other fields: an email network {Email-Enron}, a website network {Web-Stanford}, a road network dataset {RoadNet-PA}, a P2P web network dataset {P2P-Gnutella06}, and a trust network dataset {Soc-Epinions1}.

\noindent \textbf{Dataset Groups}. For conveinience of performance evaluation, we split the many datasets using different grouping methods. Firstly, two big data groups Link1 and Link2 are made using all the link prediction datasets. Notably, datasets from the same source of collection, such as Movielens-1M and Movielens-10M, or uses the same method to generate features, such as Fitness, and Photo, are put into the same group, to avoid information leakage when evaluating zero-shot performance on the other group. Apart from these two datasets, we also conduct evaluations on domain-specific groups, including E-commerce, Acadmic, and Others. Specifically, these data groups contain the following datasets, respectively: 
\begin{itemize}[leftmargin=*]
    \item \textbf{Link1}: Products-tech, Yelp2018, Yelp-textfeat, Products-home, Steam-text, Amazon-text, Amazon-book, Citation-2019, Citation-20Century, Pubmed-link, Citeseer, OGB-PPA, P2P-Gnutella06, Soc-Epinions1, Email-Enron.
    \item \textbf{Link2}: Photo, Goodreads, Fitness, Movielens-1M, Movielens10M, Gowalla, Arxiv, Arxiv-t, Cora, CS, OGB-Collab, Proteins-0, Proteins-1, Proteins-2, Proteins-3, OGB-DDI, Web-Stanford, RoadNet-PA.
    \item \textbf{Ecommerce} and \textbf{Academic}: These two datasets contain all the domain-specific datasets as mentioned above.
    \item \textbf{Others}: This group contains all the biological datasets, and other datasets including Email-Enron, Web-Stanford, RroadNet-PA, P2P-Gnutella06, Soc-Epinions1.
\end{itemize}

\subsection{Evaluation Protocols}
\label{app:protocols}
% dataset spliting
% zero-shot, and fewshot
% cross-dataset metric
All datasets used in this study are sourced from previous research as referenced~\cite{tang2024graphgpt, li2024zerog}. We adhere to the original data splits from these sources to delineate our training and testing sets. Given that many baseline methods are not equipped to manage zero-shot prediction across datasets, we instead assess their few-shot capabilities. This allows for a comparative analysis against the zero-shot performance of \model. We employ specific evaluation settings tailored to each method, detailed as follows:
\begin{itemize}[leftmargin=*]
    \item \textbf{Zero-shot Setting for \model, GraphGPT, and OpenGraph}. In our study, \model\ and two comparative graph foundation models, GraphGPT and OpenGraph, undergo evaluations for zero-shot prediction capabilities. We pre-train two instances of \model\ using Link1 and Link2 datasets. The model pre-trained on Link1 is then tested for zero-shot performance on the Link2 group datasets, and vice versa. Results labeled as "zero-shot" for \model\ are derived using this cross-evaluation method. Conversely, results marked as "full-shot" pertain to supervised learning outcomes, where, for example, the model trained on Link1 is tested on the test sets of Link1 group datasets. For GraphGPT and OpenGraph, we utilize the models as released in their respective original studies, which were pre-trained on specified datasets. \\\vspace{-0.12in}
    \item\textbf{Zero-shot Node Classification for \model}. Drawing from insights in prior research~\cite{sun2022gppt}, our approach to zero-shot node classification involves a novel method where label classes are represented as distinct nodes. We then connect existing nodes that have training labels directly to these new class nodes. This technique eliminates the need for learning specific parameters for each class within the zero-shot learning framework, streamlining the process. We have integrated this innovative approach into baseline methods as well, enhancing their capability to handle unseen node labels effectively.\\\vspace{-0.12in}
    \item \textbf{Few-shot Training for GIN and GAT}. The GIN and GAT models, employed as end-to-end training baselines, undergo training from scratch on few-shot subsets of the evaluation datasets. This approach is necessary because these models are not well-suited for cross-dataset transfer, particularly when dealing with datasets that have varying feature dimensionalities.\\\vspace{-0.12in}
    
    \item \textbf{Pre-training and Few-shot Tuning for GraphCL, GPF and GraphPrompt}. These category of baselien methods follow the pre-training-and-fine-tuning mode. In our evaluations, they are firstly pre-trained using the same pre-training datasets as our \model. Then, they experience an additional fine-tuning process using the few-shot subsets of the evaluation datasets.
\end{itemize}

\noindent \textbf{Evaluation Metrics}. For link prediction, we follow previous works~\cite{he2020lightgcn} and utilize Recall@20 and NDCG@20 as the evaluation metrics. Note that we typically use the summary results of the evaluation results across multiple datasets. Results for fifferent datasets are averaged according to their number of test samples. For the node classification task, we employ the widely-used Accuracy and Macro-F1 score as our metrics~\cite{chen2022graph, tang2024graphgpt}.

\subsection{Hyperparameter Settings}
\label{app:hyperparam}
\textbf{Optimization}. Our model, \model, is implemented using PyTorch. The optimization process employs the Adam optimizer with a learning rate of $1 \times 10^{-4}$ and a training batch size of 4096. We use cross-entropy loss with a sampled negative set~\cite{xia2023automated}. The learnable parameters of \model\ are initialized using the Xavier uniform initializer.
\textbf{Network Configurations}. The standard configuration of our \model\ includes 512 hidden units and 8 graph expert models. Each expert model comprises 8 fully-connected layers. These layers utilize a ReLU activation function and incorporate a dropout layer with a dropout probability of 0.1.
\textbf{Algorithm Hyperparameters}. The frequency regularization of our routing mechanism is set with an adjustment range of $\rho = 0.2$. The SVD decomposition is performed using 2 iterations. For structural and feature augmentation, each dataset is reprojected after using 1/10 of its samples for optimization. A minimum of 100 training steps should be executed for each dataset before its initial representations are reprojected. The reassignment of experts occurs after all training datasets have undergone one cycle of re-projection.

The baseline methods are evaluated using theeir original code or released model. We closely follow the original code to adapt to our experiments. Grid search is conducted to search for the best hyperparameter settings for each baseline method.

\subsection{Baseline Methods}
\label{app:baselines}
This section provides detailed descriptions of the baseline models used in our analysis. We employ seven different baseline models across four distinct categories.

\noindent \textbf{Training-from-scratch Graph Neural Networks}.
\begin{itemize}[leftmargin=*]
    \item \textbf{GAT}~\cite{velivckovic2017graph}. Graph Attention Networks (GAT) leverage an attention mechanism to dynamically weight node-to-node connections, enhancing the model's ability to adaptively propagate and aggregate information across the graph.\\\vspace{-0.12in}
    \item \textbf{GIN}~\cite{xu2018powerful}. The Graph Isomorphism Network (GIN) significantly boosts the expressive power of Graph Neural Networks by introducing a unique graph encoding technique aimed at effectively distinguishing between non-isomorphic graphs.\\\vspace{-0.12in}
\end{itemize}

\noindent \textbf{Graph Pre-training Models}.
\begin{itemize}[leftmargin=*]
    \item \textbf{GraphCL}~\cite{zhu2021graph}. It enhances the pre-training of graph models via self-discriminative contrastive learning, which is applied to learned node embeddings. The method employs various graph augmentation techniques such as node dropping, edge permutation, random walks, and feature masking to improve robustness.
\end{itemize}

\noindent \textbf{Graph Prompt Tuning Methods}.
\begin{itemize}[leftmargin=*]
    \item \textbf{GraphPrompt}~\cite{liu2023graphprompt}. GraphPrompt proposes a unified approach that integrates pre-training and prompt tuning for graph models. It features a learnable prompt layer designed to automatically extract crucial information from the pre-trained model to enhance performance on downstream tasks.\\\vspace{-0.12in}
    \item \textbf{GPF}~\cite{fang2022universal}. The Graph Prompt Framework (GPF) is a versatile graph prompt tuning framework compatible with various graph pre-training methods. It offers two variants of a learnable graph prompt layer, tailored to different application needs.
\end{itemize}

\subsection{Details of the Scaling Law Experiment}
\label{app:scaling_law}
For the scaling law experiment (RQ2), we elaborate the configurations of the developed instances of \model. For \model\ with different model sizes, we begin with the smallest model which has 64 hidden units, 1 fully-connected layer, and 1 expert model. The subsequent 3 model instances increases in their hidden dimensionality, from 64 to 128, 256, and 512. Then 3 larger models with more fully-connected layers are utilized, respectively containing 2, 4, and 8 MLP layers. Then we have MoE versions of \model, with 2, 4, and 8 experts, respectively. The final largest instance of \model\ has a larger latent dimensionality of 1024.

For the increase of training data, we begin with a subset of Link2 data including Cora and CS. The next version additionally includes Photo. The thir one includes ML1M. The fourth one includes Gowalla. The fifth one additionally include Arxiv and Arxiv-t. The sixth one adds the following datasets: collab, ddi, Yelp2018, Fitness, proteins-spec1, web-Stanford, proteins-spec3. The seventh one is trained with proteins-2, roadNet-PA, and Fitness additionally. And the final one is trained with all datasets from Link2.

\end{document}